\documentclass[10pt, a4paper]{article}

\usepackage[]{lrec-coling2024} 
\usepackage{multirow} 
\usepackage{booktabs}
\usepackage{array}
\usepackage[multiple]{footmisc}

\usepackage{times}
\usepackage{latexsym}

\usepackage[T1]{fontenc}

\usepackage[utf8]{inputenc}

\usepackage{microtype}

\usepackage{inconsolata}
\usepackage{fdsymbol}

\usepackage{paralist}
\usepackage{booktabs}
\usepackage{graphicx}
\usepackage[export]{adjustbox}

\newcommand{\argsubg}[2]{[#1]\textsuperscript{\textsc{#2}}}

\usepackage{verbatim}

\pagecolor[rgb]{1,1,1}
\color[rgb]{0.1,0.1,0.1}

\title{Asking and Answering Questions \\to Extract Event-Argument Structures}

\name{Md Nayem Uddin$^\spadesuit$ 
		Enfa Rose George$^\vardiamondsuit$ 
		Eduardo Blanco$^\vardiamondsuit$ 
		Steven R. Corman$^\spadesuit$} 

\address{muddin11@asu.edu, enfageorge@arizona.edu, eduardoblanco@arizona.edu, steve.corman@asu.edu\\\\
		$^\spadesuit$Arizona State University, Tempe, AZ. \\
		$^\vardiamondsuit$University of Arizona, Tuscon, AZ.  \\
         }

\abstract{
This paper presents a question-answering approach to extract document-level event-argument structures.
We automatically ask and answer questions for each argument type an event may have.
Questions are generated using manually defined templates and generative transformers.
Template-based questions are generated using predefined role-specific wh-words and event triggers from the context document. Transformer-based questions are generated using large language models trained to formulate questions based on a passage and the expected answer.
Additionally, we develop novel data augmentation strategies specialized in inter-sentential event-argument relations.
We use a simple span-swapping technique, coreference resolution, and large language models to augment the training instances.
Our approach enables transfer learning without any corpora-specific modifications and yields competitive results with the RAMS dataset.
It outperforms previous work, and it is especially beneficial to extract arguments that appear in different sentences than the event trigger.
We also present detailed quantitative and qualitative analyses shedding light on the most common errors made by our best model.
\\ \newline  }

\begin{document}

\maketitleabstract

\section{Introduction}
\label{s:introduction}
Extracting event-argument structures is an important problem in natural language  understanding~\cite{doddington-etal-2004-automatic,aguilar-etal-2014-comparison}.
At its core, it is about identifying entities participating
in events and specifying their role 
(e.g., the \emph{giver}, \emph{recipient}, and \emph{thing given} in a \emph{given} event).
Event triggers (i.e., words instantiating  events) include both
nouns~(e.g., \emph{election}, \emph{speech}),
and
verbs~(e.g., \emph{vote}, \emph{talk}).
Regardless of specific events and relations,
event-argument structures are beneficial for applications such as
news summarization~\cite{li-etal-2016-abstractive} and coreference~\cite{huang-etal-2019-improving,zhang-etal-2015-cross}. 


Traditionally, corpora 
are limited to arguments within the same sentence an event belongs to.
Inter-sentential arguments are more challenging
and have received less attention~\cite{gerber-chai-2010-beyond,ruppenhofer-etal-2010-semeval}.
Figure \ref{f:motivation} presents an example from RAMS~\cite{ebner-etal-2020-multi},
the largest corpus annotating multi-sentence event-argument structures.
Two out of four event-argument relations cross sentence boundaries.

\begin{figure}
\centering
\input{figs/motivation_tmp}
\caption{Event trigger (\emph{importing}) and its arguments in the same
 (artifact and transporter) 
  and surrounding sentences (vehicle and origin). 
  We cast the problem of extracting the arguments of an event as a question-answering task.
  Questions are automatically generated (and answered) for each argument an event may have.} 
\label{f:motivation}
\end{figure}

In this paper, we tackle the problem
of extracting event-argument structures.
As exemplified in Figure~\ref{f:motivation},
we cast the problem as a question-answering task.
We ask one question for each argument an event may have,
and rely on transformers to find answers pinpointing the text corresponding to the argument in the input document
(or, alternatively, indicate that there is no answer). 
The main contributions of this paper are:
\begin{compactitem}
\item Two approaches to formulate the questions:
  template- and transformer-based;
\item Data augmentation strategies to improve the extraction of inter-sentential arguments;
\item Quantitative results showing the benefits of our approach, including transfer learning;
and
\item Error analysis shedding light into the most challenging event-argument relations.
\end{compactitem}

The framework presented in this paper does not depend on any annotation framework, 
set of event types or argument types, domain, or corpora.
The only requirement is
a list of argument types an event may have.
Most event-argument annotation efforts satisfy this requirement,
including
PropBank~\cite{palmer-etal-2005-proposition},
NomBank~\cite{meyers-etal-2004-annotating}, 
FrameNet~\cite{baker-etal-1998-berkeley-framenet},
RAMS,
ACE~\cite{doddington-etal-2004-automatic},
and
WikiEvents~\cite{li-etal-2021-document}.
All the examples and experiments in this paper, however,
draw from the RAMS dataset~(Section~\ref{s:rams}).
We reserve for future work experimenting with other corpora.

\section{Previous Work}
\label{s:previous_work}
Extracting event-argument structures,
also referred to as event extraction~\cite{ahn-2006-stages},
includes identifying event triggers and its arguments.
The task has a long history in the field~\cite{grishman-sundheim-1996-message,doddington-etal-2004-automatic}.
Initially, datasets focused on extracting arguments within the same sentence than the verb~\cite{palmer-etal-2005-proposition,walker2006ace}.
There are also corpora focused on inter-sentential arguments~\cite{gerber-chai-2010-beyond,ruppenhofer-etal-2010-semeval,ebner-etal-2020-multi,li-etal-2021-document}.
Early models were based on handcrafted features~\cite{li-etal-2013-joint,liao-grishman-2010-using,hong-etal-2011-using}.
Like most NLP tasks, models to extract event-argument structures
experienced a transformative shift building on word embeddings, RNNs, and CNNs~\cite{chen-etal-2015-event, nguyen-etal-2016-joint-event}.

Transformer-based approaches are currently the best performing.
Some efforts assume event triggers and argument spans are part of the input and present classifiers to identify the argument type~\cite{ebner-etal-2020-multi, chen-etal-2020-joint-modeling}.
Unlike them---and like the remaining previous works discussed below---we only assume event triggers.
At a high-level, efforts to identify argument spans and argument types can be categorized into
sequence labeling, casting the problem as a question-answering task, and using generative models.
Sequence label classifiers approach the problem with the traditional BIO encoding~\cite{ramponi-etal-2020-biomedical}.
Framing the problem in terms of questions and answers is popular~\cite{du-cardie-2020-event, liu-etal-2020-event, li-etal-2020-event}.
Doing so enables zero-shot~\cite{lyu-etal-2021-zero} and few-shot~\cite{sainz-etal-2022-textual} predictions.
\newcite{li-etal-2021-document}, \newcite{ma-etal-2022-prompt} and \newcite{du-etal-2021-grit} leverage generative language models~\cite{raffel2020exploring,lewis-etal-2020-bart}.
Language generation facilitates a more flexible extraction by \emph{generating} the arguments rather than identifying spans in the input document.
Transfer learning has also been explored, including
semantic roles~\cite{zhang-etal-2022-transfer}, 
abstract meaning representations~\cite{xu-etal-2022-two}, 
and
frame-aware knowledge distillation~\cite{wei-etal-2021-trigger}.
Our approach casts the problem as a question answering task.
We introduce (a) a template- and transformer-based approach
to generate questions and (b) streamline transfer learning 
for extracting event-argument structures.

Supervised models demand annotated examples.
To mitigate this need,
unsupervised learning~\cite{huang-etal-2016-liberal, yang-etal-2018-dcfee} 
and weakly supervision~\cite{chen-etal-2017-automatically, kar-etal-2021-argfuse} have been proposed.
Data augmentation approaches~\cite{liu-etal-2021-machine, gao-etal-2022-mask} have been reported useful.
We also explore data augmentation.
Unlike previous works, our augmentation strategies target additional inter-sentential arguments.
Surprisingly, we show that arguably the simplest strategy yields the best results.


\section{The RAMS Dataset}
\label{s:rams}

Roles Across Multiple Sentences~(RAMS)~\citeplanguageresource{ebner-etal-2020-multi}
is a dataset annotating event-argument structures.
The source texts are news articles.
The annotations follow the AIDA-1 ontology.\footnote{
LDC\{2019E04, 2019E07, 2019E42, 2019E77\}}
This ontology contains a 3-level event hierarchy (e.g., \emph{transaction.transfermoney.payforservice})
and the argument types each event type may have
(e.g.,
\emph{giver},
\emph{recipient},
\emph{beneficiary},
\emph{money}, and
\emph{place}).
The ontology contains 139 events types
and
65 argument types (some are relevant to many event types, e.g., \emph{place} appears with many events).

The RAMS annotations include
(a)~event triggers (i.e., words instantiating an event type) and
(b)~the arguments of that event trigger (i.e., the word spans for each argument type).
We use the term \emph{event-argument structure} to refer to an event trigger and its arguments.
An event-argument structure need not include all the argument types in the AIDA-1 ontology (e.g., \emph{importing} is missing the \emph{destination} argument in Figure~\ref{f:motivation}).
Event triggers need not belong to the bottom level in the event hierarchy
(e.g., an event trigger may belong to \emph{transaction.transfermoney} if no child is a good fit).

The event-argument structures in RAMS are annotated across sentences.
First, annotators identified event triggers.
Second, they identified arguments~(as defined in AIDA-1) up to two sentences before or after,
as arguments are rarely found outside this window.
In practical terms, this means that documents in RAMS are 5 sentences long;
the only exceptions are event-arguments structures whose event trigger was found at the very beginning or end of the source news articles.

\begin{table}
\small
\centering
\begin{tabular}{l rrr}
\toprule 
                                & Train           & Dev        & Test \\ \midrule

\# documents                    &  3,194          & 399        & 400 \\ \addlinespace


\# events                       &  7,329          & 924        & 871\\ \addlinespace

\# arguments                    &  17,026         &2,188       &2023\\
~~~intra-sentential             &  14,018         &1,811       &1,667\\
~~~inter-sentential             &  3,008          &377         &356\\\addlinespace

\# arguments per event          &  2.23           &2.36        &2.32  \\
\bottomrule

\end{tabular}
\caption{Basic statistics of the RAMS dataset.}
\label{t:rams_stats}
\end{table}

Table \ref{t:rams_stats} presents basic statistics of the RAMS dataset. 
It includes 9,124 event-argument structures annotated in 3,993 documents.
There are 21,237 arguments in these structures.
3,741 (18\%) of these argument are inter-sentential.

\section{Asking  and Answering Questions for Event-Argument Extraction}
\label{s:approaches}
We cast the problem of extracting event-argument structures as a question answering problem and explore two
approaches:
(a) supervised learning with traditional transformers (Section~\ref{ss:approaches_supervised})
and
(b) zero- and few-shot prompts with GPT3 (Section~\ref{ss:approaches_gpt3}).
As we shall see, the latter obtains much worse results. 
Inspired by previous work~\cite{du-cardie-2020-event, liu-etal-2021-machine},
we generate questions for each argument an event may have according to the AIDA-1 ontology.
Answers are \emph{No answer} if an argument is not present.
Our novelties are as follows:
\begin{compactitem}
\item Combining two approaches to generate questions: template-based and transformer-based;
\item Exploring data augmentation strategies; and
\item Showing that our approach can easily accommodate transfer learning with other corpora.
\end{compactitem}


\subsection{Supervised Question-Answering}
\label{ss:approaches_supervised}
Using supervised learning requires us to transform the RAMS event-argument structures into a set of questions and answers.
We also explore data augmentation and transfer learning,
two strategies that can be easily incorporated into our approach.

\subsubsection{Generating Questions}
\label{ss:generating_questions}
We explore two \emph{automated} approaches to generate questions asking for the arguments of an event trigger.
The only requirement is to know a priori the name of the arguments an event may have,
an assumption we share with previous work.
Let us consider the event trigger in Figure \ref{f:motivation}, \emph{importing},
which belongs to \emph{movement.transportartifact.receiveimport}.
This event has up to five arguments in AIDA-1:
\emph{transporter}, \emph{artifact}, \emph{vehicle}, \emph{origin}, and \emph{destination}.

\paragraph{Template-Based Generation.}
We use a straightforward template to generate questions: \emph{Wh-word is the [argument type] of event [event trigger]?},
where \emph{Wh-word} is selected from the following: 
\textit{what}, \textit{where}, \textit{who}, and \textit{how}.
Questions are generated regardless of whether the argument is present in the input document.
The answer is the text corresponding to the argument if it exists
without any leading text (e.g., \emph{The answer is [argument]}, \emph{The [argument type] is [argument]}).
If the argument does not exist, the answer is \emph{No answer}.
Five question-answer pairs are generated for our running example
in the Figure \ref{f:motivation}:
\begin{compactenum}
\item Q: \emph{Who is the transporter of the event importing?}
      A: Bilal Erdogan
\item Q: \emph{What is the artifact of the event importing?}\\
      A: oil
\item Q: \emph{What is the vehicle of the event importing?}\\
      A: trucks
\item Q: \emph{Where is the origin of the event importing?}\\
      A: Syria and Iraq
\item Q: \emph{Where is the destination of the event importing?}
      A: No answer
\end{compactenum}

\paragraph{Transformer-Based Generation.}
The template-based approach results in correct question-answer pairs. 
These pairs, however, lack linguistic diversity.
In order to alleviate this issue, we experiment with T5~\cite{raffel2020exploring} to generate questions.
Specifically, we use a version of T5 pre-trained with SQuAD~\cite{rajpurkar-etal-2016-squad} to generate questions~\cite{wang-etal-2020-answer}.

The input is a document (5 sentences) and an argument.
The output 
is a 
question whose answer is the argument.
Following with the example in Figure~\ref{f:motivation}, here are the questions generated by T5:
\begin{compactenum}
\item Q: \emph{Who denied Russian allegations?}\\
      A: Bilal Erdogan
\item Q: \emph{What did Russia destroy 500 trucks with?}
      A: oil
\item Q: \emph{What type of vehicles did Russia destroy?}
      A: trucks
\item Q: \emph{Where did ISIS hold territory?}\\
      A: Syria and Iraq
\end{compactenum}

Note that T5-generated questions may be irrelevant to the event at hand.
Indeed, none of the questions above are about \emph{importing}.
Additionally, some questions are wrong. 
For example, T5 struggles with the prepositional phrase attachment in question~(2):
the \emph{oil} was carried by the \emph{trucks}---it is not what the trucks were destroyed with.
Despite these limitations,
leveraging transformer-based questions yields better results~(Section~\ref{s:results}).

There is a caveat when generating questions with T5:
we only generate questions for arguments that are present, not for all arguments in AIDA-1.
As a result, transformer-based question generation is only applicable at training time.

\subsubsection{Data Augmentation}
\label{ss:data_augmentation}
RAMS includes inter-sentential arguments, but most of them are intra-sentential.
Previous work has consistently reported worse results with inter-sentential arguments~\cite{wei-etal-2021-trigger, liu-etal-2021-machine},
so we designed six novel data augmentation strategies to improve results with inter-sentential arguments. We group the strategies into three categories: \emph{Simple Swapping, Leveraging Coreference Resolution and Leveraging LLMs.}
Figure \ref{f:data_aug_main} shows a gold example 
from the RAMS dataset and the results of
six data augmentation strategies.
We provide further details
in Appendix \ref{a:data_augmentation}.

\begin{figure*}[h!]
\centering
\input{figs/data_augmentation_main_tmp}

\caption{Examples of the data augmentation strategies
(gold event: \emph{agreements}, highlighted in red; gold argument: \emph{Clinton}, highlighted in green).
Blue highlights indicate the arguments in the augmented samples.
SS stands for Simple Swapping (P: Plain, V: Verbose),
CR for Coreference Resolution (R: Random, M: Most Meaningful),
and LLM for Large Language Model (P: Pegasus, G: GPT-3).
In the gold sample,
the event-argument is intra-sentential.
Five of the six data augmentation strategies
result in an inter-sentential argument.}
\label{f:data_aug_main}
\end{figure*}

\paragraph{Simple Swapping.}
Our first strategies are the most straightforward
and result in ungrammatical documents. 
We shift intra-sentential arguments outside their sentence and change the gold argument to point to the new position after moving to a different sentence.
We use two strategies:

\begin{compactitem}
\item \emph{Plain}. Move the intra-sentential argument into a random sentence boundary
  including the beginning and end of the document (6 options for 5-sentence documents).
\item \emph{Verbose}. Copy the intra-sentential argument into a random sentence boundary including the beginning and end of the document. We use the following template to generate the text to be pasted:
  \emph{The [argument type] of [event] is [argument]}.
\end{compactitem}

For each event-argument structure in the original training split,
each of these strategies result in as many additional event-argument structures as intra-sentential arguments in the original instance.

\paragraph{Leveraging Coreference Resolution.}
Transforming intra-sentential arguments into inter-sentential ones can be achieved by manipulating coreference chains.
We follow two strategies:

\begin{compactitem}
\item \emph{Random mention}. Update intra-sentential arguments with an inter-sentential mention
  randomly selected from its coreference chain.
\item \emph{Most Meaningful mention}. Same as \emph{Random mention} but selecting the most meaningful mention.
  We consider mentions that have more tokens and named entities (first and second criterion) to be more meaningful.
\end{compactitem}

For each event-argument structure in the original training split,
each strategy results in as many additional event-argument structures as intra-sentential arguments that
are part of a coreference chain with at least one mention belonging to another sentence.
A drawback of both strategies leveraging coreference resolution is that errors in coreference resolution lead to noisy augmented data.

\paragraph{Leveraging LLMs.}
Given the recent success of Large Language Models (LLMs),
including efforts to use them for data augmentation~\cite{yoo-etal-2021-gpt3mix-leveraging},
we also experiment with them.
Unlike the previous strategies,
using LLMs has the potential to generate unconstrained augmented data.

First, we use paraphrasing without any prompting or customization.
Specifically, we use PEGASUS~\cite{zhang2020pegasus} fine-tuned for paraphrasing~\cite{zhou-bhat-2021-paraphrase}.
Given the input document, this model returns a paraphrased version.

Second, we experiment with GPT-3 prompting~\cite{brown2020language}.
After several refinements,
we came up with the following prompt:
\emph{Rewrite the story like a newspaper article in $N$ sentences.
Include the event triggering word [event trigger]
and event arguments
[argument$_1$], [argument$_2$], [...], [argument$_n$] in the generated article.},
where $N$ is the number of sentences in the original document. 

Rewriting the input document requires us to modify the gold span positions.
The process is conceptually simple,
but neither PEGASUS nor GPT3 are guaranteed to keep the tokens for each argument in the generated text.
The mapping process inevitably results in unmapped events and argument.
Out of 7,079 events and 17,026 arguments,
we successfully map
66.6\% events and 74.8\% arguments with PEGASUS
and
90.2\% events and 93.5\% arguments with GPT3.

\subsubsection{Transfer Learning}
\label{ss:transfer_learning}
Transfer learning has been shown useful for extracting event-argument structures among others.
Previous efforts project annotations~\cite{huang-etal-2018-zero}
or reuse existing corpora in a specialized manner~\cite{zhang-etal-2022-transfer}.
We take a streamlined approach:
transform existing corpora into questions and answers using the \emph{same} methods described in Section \ref{ss:generating_questions}.
We work with:

\noindent
\textbf{ACE}~\citeplanguageresource{walker2006ace}.
It contains 5,349 event triggers annotated in broadcast conversations and news, newsgroups, phone conversations, and weblogs.
ACE considers 8 event types, 33 event subtypes, and 36 argument types.

\noindent
\textbf{WikiEvents}~\citeplanguageresource{li-etal-2021-document}.
It contains 3,951 event triggers annotated in Wikipedia pages.
It includes 50 event types and 59 argument types.

\noindent
\textbf{QA-SRL}~\citeplanguageresource{he-etal-2015-question}.
It includes crowdsourced questions and answers encoding predicate-argument structures.
Argument do not have a specific type; the question wording captures their role.
QA-SRL includes 299,308 question-answer pairs. 


\begin{table*}
\small
\centering
\begin{tabular}{l rr}
\toprule
 & \multicolumn{1}{c}{Base} & \multicolumn{1}{c}{Large} \\ \midrule

Supervised (RAMS) with Template-Based Questions     & 42.58\scriptsize{\(\pm\)0.73}                                         & 48.23\scriptsize{\(\pm\)0.82} \\
~~~+ Merging Transformer-based Questions            & 45.39\scriptsize{\(\pm\)0.13}\makebox[0pt][l]{$^{*}$}                 & \bf 50.69\scriptsize{\(\pm\)1.52}\makebox[0pt][l]{$^{*}$}  \\
~~~~~~+ Blending Augmented Data \\
~~~~~~~~~~Simple Swapping \\
~~~~~~~~~~~~~Plain (\(\alpha = 0.4\))               & 45.20\scriptsize{\(\pm\)0.75}\makebox[0pt][l]{$^{* \dagger}$}         & 49.61\scriptsize{\(\pm\)0.93}\makebox[0pt][l]{$^{*}$}  \\
~~~~~~~~~~~~~Verbose (\(\alpha = 0.6\))             & \bf 46.86\scriptsize{\(\pm\)0.30}\makebox[0pt][l]{$^{* \dagger}$}     & 49.97\scriptsize{\(\pm\)0.41}\makebox[0pt][l]{$^{*}$}  \\
~~~~~~~~~~Leveraging Coreference Resolution \\
~~~~~~~~~~~~~Random mention (\(\alpha = 0.4\))      & 44.65\scriptsize{\(\pm\)0.92}\makebox[0pt][l]{$^{*}$}                 & 48.38\scriptsize{\(\pm\)0.52}\makebox[0pt][l]{$^{*}$}  \\
~~~~~~~~~~~~~Meaningful mention (\(\alpha = 0.4\))  & 45.89\scriptsize{\(\pm\)0.80}\makebox[0pt][l]{$^{* \dagger}$}         & 49.12\scriptsize{\(\pm\)0.55}\makebox[0pt][l]{$^{* \dagger}$}  \\
~~~~~~~~~~Leveraging Large Language Models \\
~~~~~~~~~~~~~Pegasus    (\(\alpha = 0.2\))          & 46.72\scriptsize{\(\pm\)1.11}\makebox[0pt][l]{$^{* \dagger}$}         & 47.72\scriptsize{\(\pm\)0.21} \\
~~~~~~~~~~~~~GPT3         (\(\alpha = 0.4\))        & 45.89\scriptsize{\(\pm\)1.55}\makebox[0pt][l]{$^{* \dagger}$}         & 48.45\scriptsize{\(\pm\)0.37}\makebox[0pt][l]{$^{*}$}  \\ 

 \bottomrule
\end{tabular}

\caption{Results (F1) obtained with the test split of RAMS (mean and standard deviation of five runs).
  Merging transformer-based questions is useful with both base and large models whereas blending augmented data yields improvements with base models.
  We indicate statistically significantly better results (McNemar test \cite{McNemar1947}, $p<0.01$)
  with respect to \emph{Supervised (RAMS) with Template-Based Questions} with an asterisk ($^{*}$), and to \emph{Merging Transformer-based Questions} with a dagger ($^{\dagger}$).}
\label{t:results}
\end{table*}

\subsection{Zero- and Few-Shot Question Answering}
\label{ss:approaches_gpt3}
Large language models are credited with having emergent abilities~\cite{wei2022emergent}.
They are also capable of in-context learning~\cite{brown2020language},
meaning that they can (presumably) solve problems with a small number of training examples when given instructions~\cite{wang-etal-2022-super}.

In order to test the aforementioned abilities when it comes to extracting event-argument structures across sentences,
we experiment with GPT-3 and zero- and few-shot approaches. 
We provide prompt examples with details in Appendix \ref{a:gpt3}

\noindent
\textbf{Zero-Shot.}
We prompt GPT-3 with the input document (five sentences) and the questions generated with the template-based approach~(Section~\ref{ss:generating_questions}).
Note that the transformer-based approach to generate questions cannot be used as it requires the answers to the questions
(i.e., the arguments we are prompting GPT-3 to find).

\noindent
\textbf{Few-Shot.}
Few-shot prompts is similar to zero-shot prompts except that they are preceded by two randomly selected data samples from the training split
(using the same format than the expected answer).
These examples also include questions without answers.

\section{Quantitative Results and Analyses}
\label{s:results}

We present results with the test split of RAMS (mean and standard deviation of five runs).\footnote{Code including dataset transformed into questions and answers are available at \href{https://github.com/nurakib/event-question-answering}{https://github.com/nurakib/event-question-answering}}
All our models are tuned with the train and validation splits of RAMS
(and the same splits of the additional corpora with transfer learning).
For data augmentation with coreference, we use the model by~\newcite{clark-manning-2016-deep}.
All our models use RoBERTa~\cite{liu2019roberta} for extractive question answering, similar to~\cite{du-cardie-2020-event}.
We conducted experiments utilizing the base and large variants of RoBERTa 
to assess performance relative to the model size. 
We use Pytorch~\cite{paszke2019pytorch} and HuggingFace transformers~\cite{wolf-etal-2020-transformers}.
The only exception is GPT-3, which has its own API.

All models are evaluated using the official RAMS evaluation script (P, R, and F1; exact match with the ground truth).
The only exception is GPT-3, with which we use a more lenient evaluation.
Since the input document sometimes contains the text generated by GPT-3 more than once,
there are multiple options to map the GPT-3 generated answer to token indices.
We consider GPT-3 is correct if any of the mappings matches the ground truth.


\paragraph{Merging and Blending}
Using transformer-based questions, data augmentation, and transfer learning requires us to decide how to leverage
these additional instances during the training process.
Note that transformer-based questions are new question-answer pairs derived from the original RAMS instances,
whereas data augmentation and transfer learning increase the number of training instances---%
and the question-answer pairs.

We explore two options: merging and blending.
Merging is the simpler option: concatenate all the question-answer pairs and consider them equal during the training process.
Blending~\cite{shnarch-etal-2018-will} is more complicated and relies on the intuition that some instances
(in our case, the question-answer pairs generated from RAMS) ought to be given more importance than the additional training instances.
Specifically, blending starts the training process (first epoch) with the result of merging RAMS and additional instances.
Then, in each epoch, the amount of additional instances is reduced by a hyperparameter $\alpha$.

\paragraph{Results training with RAMS}
Table \ref{t:results} presents the results with RoBERTa base and large models.
We experiment with both merging and blending transformer-based questions and augmented data samples,
resulting in four distinct combinations.
Among these, we identified the most effective approach—merging transformer-based questions followed by blending augmented samples—which is presented in Table \ref{t:results}.
Merging transformer-based questions brings a 2.81 F1  improvement (42.58 vs. 45.39) with the base model
and a 2.46 F1 improvement (48.23 vs 50.69) with the large model.
This is the case despite these automatically generated questions are noisy and are often worded with respect to other events than the event of interest (Section~\ref{ss:generating_questions}).
Blending augmented data samples (\emph{Simple Swapping}) is also beneficial with the base model (+1.47 F1; 45.39 vs. 46.86).
Surprisingly, data augmentation strategies are not beneficial with the large model. 

To summarize the results of merging and blending in Table \ref{t:results},
(a)~merging transformer-based questions is more beneficial than blending,
and
(b)~blending augmented data is more beneficial than merging.
These results are intuitive,
as transformer-based questions are rarely nonsensical but data augmentation is noisy---simple swapping results in non-grammatical texts, coreference resolution is noisy, 
and LLMs rephrasing is non-deterministic.
In other words, blending outperforms merging when the additional data is less reliable.

Transfer learning brings statistically significantly better results (Table \ref{t:results_transfer})
with the base (F1: 48.53 vs. 46.86) and large model (52.89 vs. 50.69).
QA-SRL, unlike ACE and WikiEvents,
does not annotate explicit event-argument structures.
Instead, it encodes them in the wording of questions.
We hypothesize that QA-SRL yields worse results because
questions in QA-SRL are actual natural language written by crowdworkers rather than
the result of instantiating templates or T5.

\paragraph{Comparison with Previous Work.}
Our best model outperforms the best published results with RAMS (Table \ref{t:results_transfer}):
PAIE~\cite{ma-etal-2022-prompt} obtains 52.2 F1 and we obtain 52.89 F1 (both large).
Using the base model, however, we do not outperform previous work: 
PAIE obtains 49.5 F1 and we obtain 48.53 F1.
We point out that
PAIE requires role-specific parameters, meaning that unlike our approach, it cannot easily accommodate transfer learning and it is unable to make zero-shot predictions.
Additionally, PAIE uses BART as the underlying pre-trained model, which has 15\% more parameters than the one we use, RoBERTa.

\begin{table}
\small
\centering
\begin{tabular}{l lr@{}lr}
\toprule
                            & \multicolumn{1}{c}{PLM} & \multicolumn{2}{c}{Base} & \multicolumn{1}{c}{Large} \\ \midrule

RAMS                        & RoBERTa & 46.86&\scriptsize{\(\pm\)0.30}                                & 50.69\scriptsize{\(\pm\)1.52} \\
+ACE                        & RoBERTa & \bf 48.53&\scriptsize{\(\pm\)1.30}\makebox[0pt][l]{$^{*}$}    & \bf 52.89\scriptsize{\(\pm\)0.61}\makebox[0pt][l]{$^{*}$} \\
+QA-SRL                     & RoBERTa & 44.94&\scriptsize{\(\pm\)0.79}\makebox[0pt][l]{$^{*}$}        & 44.90\scriptsize{\(\pm\)1.05}\makebox[0pt][l]{$^{*}$} \\
+WikiEvnt                   & RoBERTa & 46.58&\scriptsize{\(\pm\)0.68}                                & 51.46\scriptsize{\(\pm\)0.80} \\ \midrule

\multicolumn{3}{l}{Previous Work (top 2)} \\
~~~PAIE                     & BART    & 49.50&\scriptsize{\(\pm\)0.65}                                & 52.20\scriptsize{\(\pm\)n/a} \\
~~~TSAR                     & BART    & 48.06&\scriptsize{\(\pm\)n/a}                                 & 51.18\scriptsize{\(\pm\)n/a} \\ \bottomrule

\end{tabular}

\caption{Results (F1) obtained merging RAMS and related corpora (mean and standard deviation of five runs).
  We retrain the best model using only RAMS (boldfaced in Table \ref{t:results}, same as first row) and
  indicate statistically significantly better results (McNemar test~\cite{McNemar1947}, $p<0.01$)
  with an asterisk ($^{*}$).
  We compare with 
  PAIE~\cite{ma-etal-2022-prompt} and TSAR~\cite{xu-etal-2022-two}.
  }
\label{t:results_transfer}
\end{table}

\paragraph{Comparison with GPT-3}
Zero-shot and few-shot prompting with GPT-3 obtains much worse results
than our supervised question-answering approach.\footnote{We only report one run for zero- and few-shot with GPT-3 as we do not train it.}
The results in Table \ref{t:gpt3_comparision} show that GPT-3 obtains better results 
in a few-shot in-context learning setting, yet the performance lags behind the supervised models by a significant margin.

\begin{table}
\small
\centering
\begin{tabular}{l rrr}
\toprule
                            & \multicolumn{1}{c}{P}           & \multicolumn{1}{c}{R}            & \multicolumn{1}{c}{F1} \\ \midrule

Ours, base            & 54.48\scriptsize{\(\pm\)0.86}             & 43.47\scriptsize{\(\pm\)1.76}       & 48.53\scriptsize{\(\pm\)1.30} \\
Ours, large           & 60.90\scriptsize{\(\pm\)0.46}             & 46.74\scriptsize{\(\pm\)0.78}       & 52.89\scriptsize{\(\pm\)0.61} \\ \midrule

\multicolumn{3}{l}{GPT-3} \\
~~~Zero-shot               & 27.3                & 21.4       & 24.0 \\
~~~Few-shot                & 32.6                & 29.1       & 30.7 \\\bottomrule

\end{tabular}
\caption{Results obtained with our models (boldfaced in Table~\ref{t:results_transfer}) and GPT-3 (\emph{text-davinci-003}).}
\label{t:gpt3_comparision}
\end{table}

\paragraph{Are Inter-Sentential Arguments Harder?}
Table \ref{t:results_distance} details the results of our best models (boldfaced in Table~\ref{t:results_transfer})
broken down by the number of sentences between the event trigger and argument.
The improvements (\%$\Delta$F1) with respect to the simplest question-answering approach
(only template-based questions, no data augmentation and no transfer learning; first supervised model in Table~\ref{t:results})
are substantial regardless of distance between the event trigger and argument.
For the base model,
we observe 402\% improvement when the argument appears two sentences after the event trigger.
The improvements are substantial 
when arguments appear in the sentences before (36.5\% F1, 30.5\% F1) or the sentence after (45.6\% F1).
For the large model,
arguments in the sentence before the event benefit the most (44.6\% F1),
followed by those two sentences before (38.5\% F1).
Arguments after the event also benefit (6.5\% and 18.2\% F1).
In summary, our model is beneficial regardless of where the argument appears with respect to the event.

\begin{table}
\small
\centering
\begin{tabular}{l rrrr}
    \toprule
    \multirow{2}{*}{} & \multicolumn{2}{c}{Base} & \multicolumn{2}{c}{Large}\\ \cmidrule(lr){2-3} \cmidrule(lr){4-5}
     & \multicolumn{1}{c}{F1} & \multicolumn{1}{c}{\%$\Delta$F1} & \multicolumn{1}{c}{F1} & \multicolumn{1}{c}{\%$\Delta$F1} \\
    \midrule
    2 before     & 29.17      & +36.5     & 30.30    & +38.5 \\
    1 before     & 31.23      & +30.5     & 34.56    & +44.6 \\
    same         & 54.43      & +12.5     & 56.98    &  +6.7 \\
    1 after      & 22.08      & +45.6     & 22.76    &  +6.5 \\
    2 after      & 27.91      & +402.0     & 22.73    & +18.2 \\
    \bottomrule
\end{tabular}
\caption{Results by our best models (boldfaced in Table~\ref{t:results_transfer})
  broken down by distance (\# sentences) between arguments and events.
  \%$\Delta$F1 indicates the relative improvement with respect to training only with template-based questions and RAMS (first supervised model in Table \ref{t:results}).
  Our approach benefits all arguments, especially those that are not in the same sentence than the event.}
\label{t:results_distance}
\end{table}

\begin{figure}
\includegraphics[width=\columnwidth]{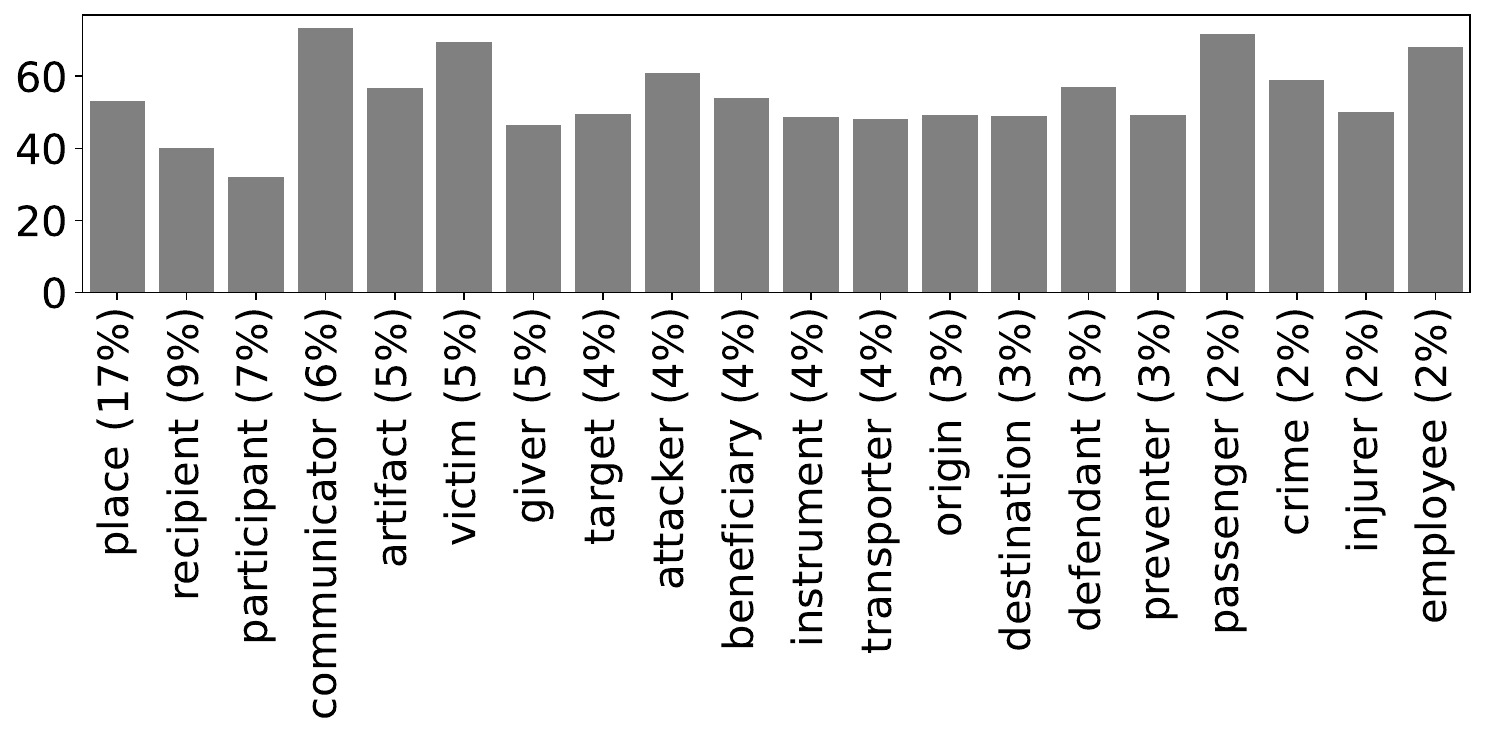}
\caption{F1 per argument of our best model (boldfaced in Table \ref{t:results_transfer}, large).
  Frequency in training (between parenthesis) is only a weak indicator of F1,
  leading to the conclusion that some arguments are easier to learn.
  For example, \emph{employee} is less frequent than \emph{participant} yet the former obtains twice the F1 (0.70 vs. 0.33).}
\label{f:results_args}
\end{figure}

\paragraph{Are Frequent Arguments Easier?}
It is a common belief that the more training data the easier it is to learn.
Figure \ref{f:results_args} provides empirical evidence showing that this is not the case when predicting event-argument structures in RAMS.
We observe that per-argument F1 scores range from 0.34 to 0.70,
but there is no pattern indicating that frequency correlates with F1 score.
For example, infrequent events such as \emph{employee} and \emph{passenger} (2\%) obtain results
as high as those obtained with communicator (6\%) and victim (5\%).

\begin{figure}
\includegraphics[width=\columnwidth]{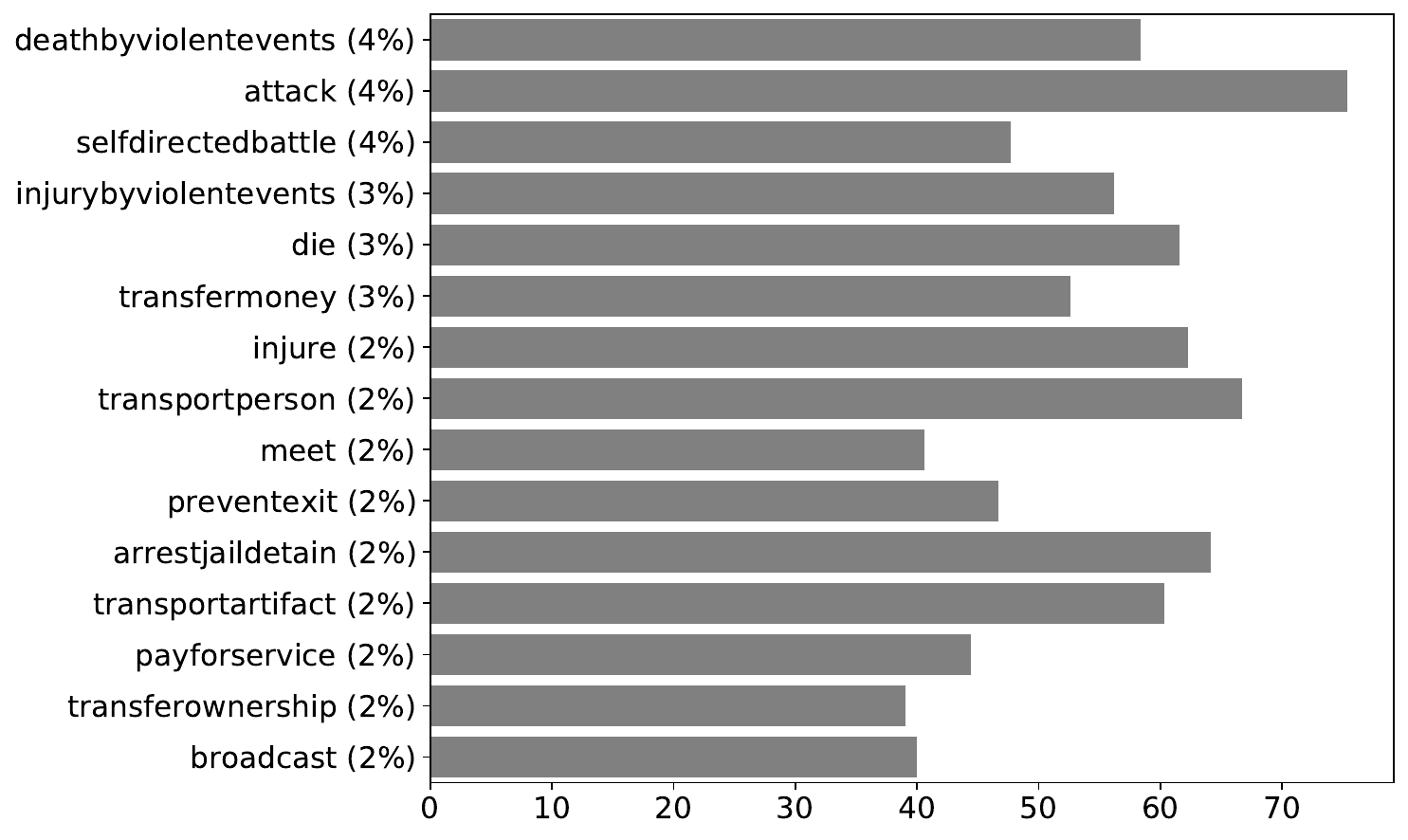}
\caption{Average F1 per event (top 15 most frequent events)
  by our best model (boldfaced in Table \ref{t:results_transfer}, large).
  There is no clear relation between event frequency in training (between parenthesis) and F1,
  leading to the conclusion that arguments of some events are easier to learn (e.g., \emph{selfdirectedbattle} vs. \emph{transfortartifact})}
\label{f:results_events}
\end{figure}

\paragraph{Are Frequent Events Easier?}
No, they are not.
Surprisingly, more training data for an event does not always lead to better results.
Figure \ref{f:results_events} shows the average F1 for the top 15 most frequent events.
The graph shows no clear pattern between event frequency in training and F1.
Indeed, arguments of events with 2\% frequency obtains F1-scores ranging from 0.33 and 0.60,
a large range that overlaps with the F1-scores of more frequent events.

\begin{figure}
\includegraphics[width=\columnwidth]{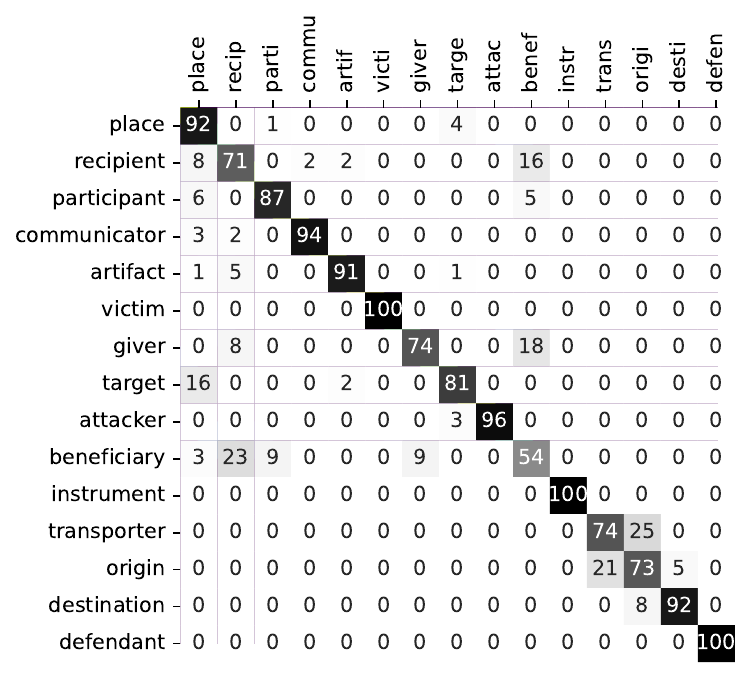}
\caption{Confusion matrix comparing gold (rows) and predicted (columns) argument types 
  for correctly predicted argument spans (top 15 most frequent types).
  Most errors are plausible (at face value) but semantically wrong argument types
  (e.g., mislabeling the \emph{beneficiary} as the \emph{recipient}; note that both are usually people).} 
\label{f:args_mislabeled}
\end{figure}

\paragraph{Which Arguments are Mislabeled?} 
Our best model obtains 52.89 F1.
This number is low, but the evaluation is strict:
it expects predictions to match the exact text span and argument type.
Figure \ref{f:args_mislabeled} compares gold (rows) and predicted (columns) argument types
when our best model (boldfaced in Table \ref{t:results_transfer}, large) predicts the correct argument spans.
We observe two main trends.
First, the model mislabels arguments with labels that are plausible at first sight but wrong given the input document.
For example, \emph{recipient}, \emph{beneficiary}, and \emph{giver} are often people but they have different semantics given an event trigger in context.
Second, our model mislabels arguments that could be considered correct but do not follow the RAMS annotations. 
For example, the \emph{transporter} of a \emph{transporting} event~(i.e., the person moving something)
could be the \emph{origin} or the event,
but RAMS uses that argument type for the location where \emph{transporting} started.
We hypothesize that our model leverages the knowledge acquired about \emph{transporter} and \emph{origin}
prior to our training with RAMS (and it never overcame this prior knowledge).

\begin{table*}[htp!]
\small
\definecolor{goldenbrown}{rgb}{0.6, 0.4, 0.08}

\begin{tabular}{p{0.75in} p{5.25in}}
\toprule
Error Type & Example \\ \midrule

Partial spans (38.51\%) &
The Trump Wall, the past shows, does not promise a solution to the forces driving migration along  {\textcolor{blue}{[the}} { \textcolor{goldenbrown} {\argsubg{U.S.-Mexico border.}{gold\_place}}}{\textcolor{blue}{]\textsuperscript{\textsc{predicted\_place}}}} But it does offer the illusion of a solution. So if the Trump Wall is ever built, no one should be surprised when it is bypassed, breached or {\argsubg{bombarded}{event\_trigger}}, just like those that came before it. \\ \addlinespace

Alternatives (13.51\%) & [\dots] Then she gave an expansive denunciation of  {\textcolor{blue}{ \argsubg{Pakistan}{predicted\_jailer}}}. Since its creation, { \textcolor{goldenbrown} {\argsubg{it}{gold\_jailer}}} had \argsubg{jailed}{event\_trigger} or exiled rival politicians. [\dots]
 \\ \addlinespace


Distractors (4.05\%) & 
[\ldots] has published
documents such as the probable-cause affidavit in a lieutenant's pain-pill addiction case, \argsubg{purchase}{event\_trigger} orders showing that the { \textcolor{goldenbrown} {\argsubg{sheriff's office}{gold\_giver}}} spent more than \$60,000 
[\ldots] 
Now a technology consultant who regularly travels to Russia, {\textcolor{blue}{ \argsubg{Dougan}{predicted\_giver}}} says he made friends with hackers there and \underline{sold his website to them}.
\\  \addlinespace

Wrong spans (6.76\%) & [\ldots] As a result, for the second time in four months the ratings agency S\&P has downgraded Saudi Arabia's debt rating, which makes it more expensive for Saudi Arabia to borrow {\textcolor{blue}{ \argsubg{money}{predicted\_recipient}}}. The { \textcolor{goldenbrown} {\argsubg{country}{gold\_recipient}}} is reportedly also asking banks for a \argsubg{loan}{event\_trigger} of up to \$10 billion (\pounds{}6.8 billion) [\ldots] \\ \addlinespace

Two or more arguments (2.70\%) &
On Wednesday’s Breitbart News Daily, Sirius XM host { \textcolor{goldenbrown} {\argsubg{Alex Marlow}{gold\_participant\_1}}} {\textcolor{blue}{ $\textsuperscript{predicted\_participant}$}} \argsubg{discussed}{event\_trigger} leaked Hillary Clinton emails with { \textcolor{goldenbrown}{\argsubg{former Navy SEAL and Blackwater CEO Erik Prince.}{gold\_participant\_2}}}\\  \bottomrule

\end{tabular}

\caption{Most common errors made by our best performing model 
(boldfaced in Table \ref{t:results_transfer}, large).}
\label{t:errors}
\end{table*}

\section{Error Analysis}
\label{s:error_analysis}

We close the analyses examining the errors made by the best system (boldfaced in Table \ref{t:results_transfer}, large).
We discuss linguistic commonalities in either the input documents or system predictions
observed in a manual analysis of all the errors made in 100 documents (148 errors).

The majority of errors (38.51\%, Table \ref{t:errors}) are due to predicting \textit{partial spans} (either shorter or longer than the gold).
The differences include articles, conjunctions, numbers, and detailed descriptions complementing entities.
Completely \textit{wrong spans} are much less likely (6.76\%).
Despite we could not identify the underlying cause of all \emph{wrong spans},
there are two common causes.
First, the ground truth includes \emph{one token span} per argument, but valid \emph{alternatives} are sometimes present (13.51\% of errors).
In the example, our system predicts a coreferent mention that is counted as an error.
Second, \emph{distractors} sometimes mislead the system.
In the example, the system appears to confuse the event trigger (i.e., purchase) with a semantically similar but unrelated event: \emph{sold his website}.

Our system is limited to predicting one span per argument type,
thus it will always make errors with events that have two instances of the same argument type (2.70\% of errors).
A previously reported by~\newcite{zhang-etal-2020-two},
we found that some errors (6.08\%) appear to be due to annotation errors---no annotations are perfect, and RAMS is not an exception.
For example, the test set includes the following: 
\textit{he raised the \argsubg{funds}{ recipient} privately and \argsubg{reimbursed}{ event\_trigger} the city [\ldots]}. 

\section{Conclusions}
\label{s:conclusions}

We have presented an approach to extract event-argument structures by automatically asking and answering questions.
Our approach combines two complementary strategies to generate questions:~%
template- and transformer-based.
The latter not only generates noisy question-answer pairs,
but also correct pairs involving events other than the event of interest.
Yet, using transformer-based questions yields better results.
Further, we explore several data augmentation strategies targeting inter-sentential arguments,
as they are harder to identify.
Transforming intra-sentential arguments into inter-sentential arguments by moving them to random sentence boundaries
is the best strategy when experimenting with RoBERTa base.
Indeed, it yields better results than leveraging coreference resolution or large language models
(and compounding their errors).

Our transformer-based question generation combined with transfer learning outperforms previous work with RoBERTa large. 
Also, the data augmentation strategies helped the base model to achieve better or comparable results to (i.e., within a standard deviation)
the top-2 best performing previous works.
We use 11-14\% less parameters, and, crucially, our model does not have any role-specific parameters.
The lack of role-specific parameters has two advantages.
It allows us to 
streamline transfer learning
and
make zero-shot predictions.

\section{Limitations}
The work presented in this paper has several limitations.
Our model is limited to predict only one argument type per event trigger.
Thus, any even trigger that has two arguments with the same argument type is guaranteed to yield an error (Section \ref{s:error_analysis}). Addressing this limitation requires further work, including multi-turn question answering. 

Despite the empirical benefits of transformer-based questions, 
they are noisy and potentially nonsensical (Section \ref{ss:generating_questions}). 
Our model learns from the noisy questions, 
but further work is needed to understand why and improve the generation process. 
We did a small-scale analysis but could not identify recurrent errors to improve the transformer-based question generation.
While our best model shows improvement across all event-argument relations, the benefits are most noticeable in inter-sentential arguments. That said, results are still better with intra-sentential arguments. Further work is needed to extract event-argument structures from long documents.

\newpage

\section{Acknowledgments}
This research was supported by a grant from the U.S. Office of Naval Research (N00014-22-1-2596).
We would like to thank the anonymous reviewers for their
insightful comments and suggestions.

\section{Bibliographical References}
\label{sec:reference}
\bibliographystyle{lrec-coling2024-natbib}
\bibliography{lrec-coling2024}

\section{Language Resource References}
\label{lr:ref}
\bibliographystylelanguageresource{lrec-coling2024-natbib}
\bibliographylanguageresource{languageresource}

\appendix

\section{Zero- and Few-Shot Prompts}
\label{a:gpt3}

To conduct our experiments with GPT-3 in zero and few-shot settings,
we utilized the OpenAI API.\footnote{\href{https://openai.com/}{OpenAI}}
Specifically, we use the `text-davinci-003' model and the `Completion' endpoint provided by the API. 

To ensure consistency with the inputs used in the supervised model, we designed the prompts for the GPT-3 model in a similar manner. However, there were slight differences in how the prompts were handled between the supervised settings and the zero and few-shot experiments. In the supervised settings, we posed one question per iteration, whereas, for the GPT-3 zero-shot and few-shot experiments, we included all the questions simultaneously. We conducted a small-scale study using a subset of samples from the RAMS dataset to validate the impact of asking all questions at once compared to asking one question per iteration. Our study did not reveal any difference between the two approaches. Hence, we proceeded with asking all questions at once for the zero and few-shot experiments. This streamlined the experimentation process and also helped to reduce the costs of querying the API. 

\subsection{Example of Zero-Shot Prompt}

\begin{figure}
\centering
\input{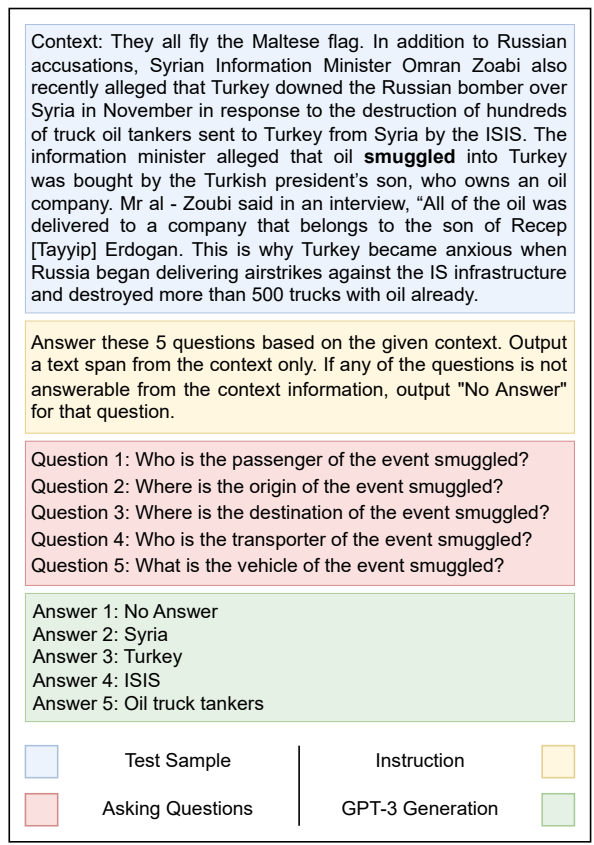}
\caption{Example of a zero-shot GPT-3 prompt. \emph{Test sample}, \emph{Instruction} and \emph{Asking Questions} are all together considered the prompt (\emph{input}) and \emph{GPT-3 Generation} is the output.}
\label{f:zeroshot}
\end{figure}

In the zero-shot setting, our objective is to extract event arguments without any training examples. To accomplish this, we construct prompts with template-based questions. The GPT-3 model generates answers to these questions, which are then mapped back to the provided document to extract matched event argument spans. Figure \ref{f:zeroshot} presents a snapshot of a zero-shot prompt. 

\subsection{Example of Few-Shot Prompt}

\begin{figure*}[h!]
\centering
\input{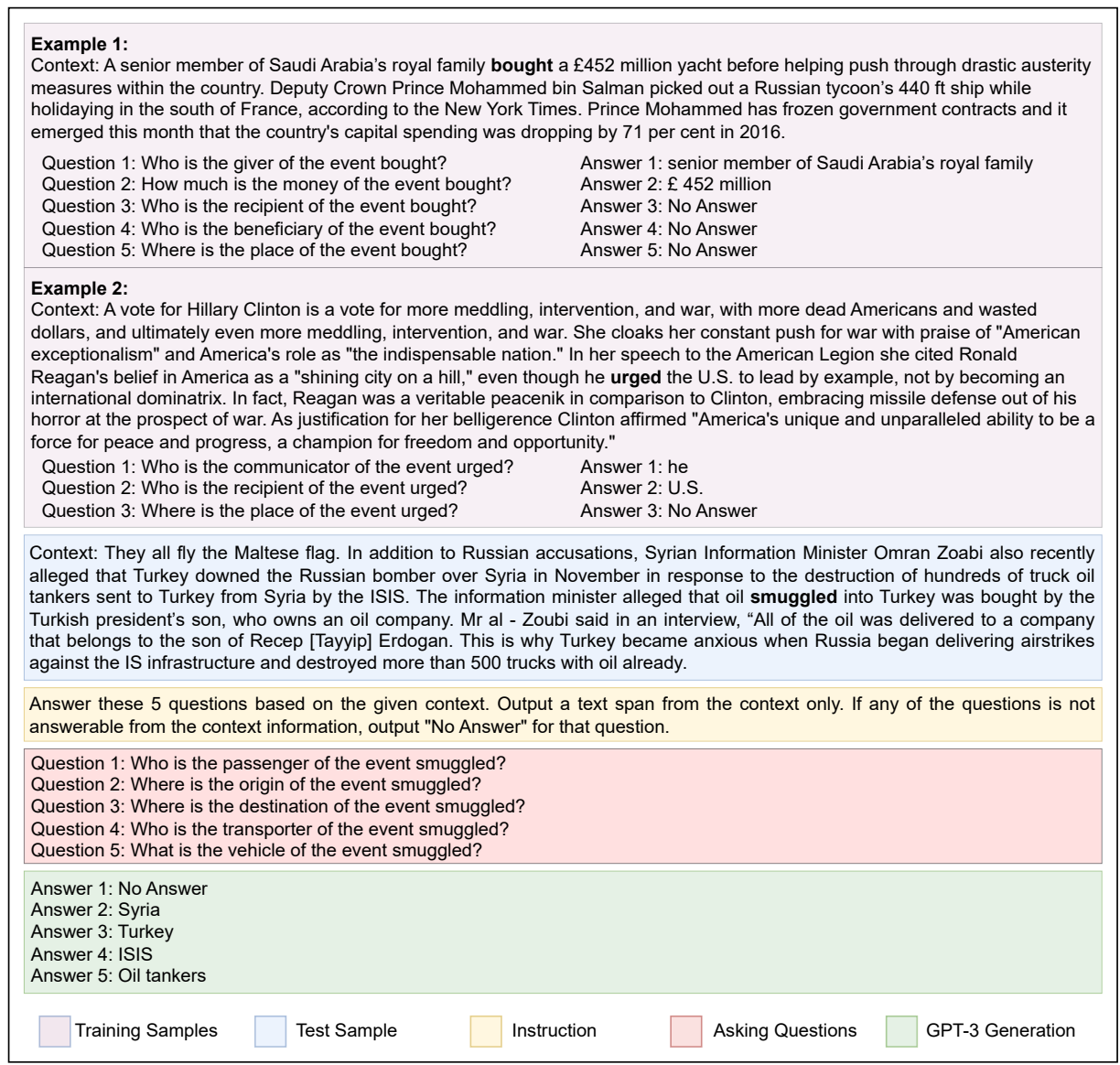}

\caption{Example of a few-shot GPT-3 prompt. \emph{Training samples}, \emph{Test sample}, \emph{Instruction} and \emph{Asking Questions} are all together considered as prompt (\emph{input}) and \emph{GPT-3 generation} is the output.}
\label{f:fewshot}
\end{figure*}

In the few-shot setting, we leverage a limited amount of training data. We randomly select two training samples from the RAMS dataset. These examples are formatted to match the inputs used during supervised training. By incorporating two training samples, we enhance the model's ability to capture event arguments and generate accurate responses. Figure \ref{f:fewshot} presents a snapshot of a few-shot prompt. 

\section{QA Model and Hyperparameters}
\label{a:hyperparameters}

In this section, we provide an overview of the model and hyperparameters used in the supervised approach for event argument extraction. We leveraged the RoBERTa-base and large models to generate contextual representations for the question and document pairs. By feeding the question and document as input to the RoBERTa model, we obtained the contextualized representation of the combined text. Then, we employ a task-specific layer that operates on top of these representations. This layer is responsible for predicting the start and end offsets of the argument span.

During training, we utilize annotated data samples where the ground truth start and end offsets of the argument span are provided. The output layer is trained using the cross-entropy loss function to minimize the discrepancy between the predicted offsets and the ground truth offsets. We conducted experiments using three different learning rates [2e-5, 3e-5, 5e-5] and dropout values [0.3, 0.4, 0.5] to optimize the performance of the models. In order to determine the optimal hyperparameters, we evaluated their performance on the validation dataset. This evaluation allowed us to select the final hyperparameters that yielded the best results. We list all the hyperparameters in Table \ref{t:parameters}. 

To mitigate the risk of overfitting and ensure efficient training, we incorporated the technique of early stopping. If the loss function fails to show  improvement over 10 consecutive epochs, training is terminated before completing 200 epochs. 

At inference time, given a new question and document pair, the trained model applies the learned weights and biases to predict the most likely start and end offsets of the argument span. These predicted offsets indicate the span within the document where the argument is expected to be found.

\begin{table}
\centering
\begin{tabular}{l r}
\toprule 
Name                            & Value           \\ \midrule

learning rate                   &  2e-5, 3e-5, 5e-5          \\ 
number of epochs                &  200          \\ 
patience 			     		&  10   		\\
dropout                         &  0.3, 0.4, 0.5 \\
training batch size             &  8       \\
validation batch size           &  4         \\
test batch size                 &  4          \\
max length                      &  512          \\
loss function                   &  cross-entropy  \\
optimizer                       &  Adam          \\
blending factor ($\alpha$)      &  0.2, 0.4, 0.6  \\ \bottomrule

\end{tabular}
\caption{Hyperparameters of the supervised models trained on RAMS.}
\label{t:parameters}
\end{table}
\section{Data Augmentation Examples}
\label{a:data_augmentation}
In this section, we demonstrate the three data augmentation strategies. The goal of these data augmentation strategies is to transform intra-sentential arguments into inter-sentential arguments while generating new training instances. For demonstration purposes, we select a data sample from RAMS, featuring the event triggering word \emph{agreement} and two event arguments: \emph{Clinton}~(violator) and \emph{Iran}~(otherparticipant).

\subsection{Simple Swapping}
The simple swapping strategy involves shifting an intra-sentential argument to different positions (at the beginning or end position of each sentence), transforming it into an inter-sentential argument. The given example in Table \ref{t:augmentation_examples_1} has 5 sentences, so that leaves 6 different places to shift the argument.
We consider the end position of the $i$th sentence and the beginning position of the $(i+1)$th sentence as the same position. Then, one position is chosen randomly from these 5 positions to replace the corresponding gold annotation. The \textit{verbose} version of the simple swapping approach follows the same procedure for determining the new position of the argument. However, we replace the argument with a simple sentence such as \emph{``The violator of the event agreement is Clinton.''} Also, we keep the original and the augmented argument in the document whereas we discard the original argument for the simple swapping (\emph{Plain}). It is worth noting that both versions generate grammatically incorrect sentences, but we focused on generating argument spans that are inter-sentential.

\subsection{Leveraging Coreference Resolution}
In the coreference-based data augmentation strategy, the first step involves identifying the coreference chains related to the given argument. In Table \ref{t:augmentation_examples_1}, the sample exhibits two coreference chains corresponding to the arguments. These chains are extracted using the spaCy library.\footnote{\href{https://spacy.io/}{spaCy}} For the argument \emph{Clinton}, the coreference chain appears as \emph{Hillary Clinton: [Hillary Clinton (sent 1), she (sent 2), Clinton (sent 3)]}. Similarly, for the argument \emph{Iran}, the coreference chain is \emph{[Iran (sent 2), Iran (sent 3), its (sent 3), the country (sent 3), Iran (sent 4)].} To generate augmented data using these coreference chains, we randomly select a mention from the coreference chain to replace the corresponding gold annotation. Alternatively, for the \textit{most meaningful} mention, we prioritize the selection of the argument with the highest number of tokens and named entities, such as choosing \emph{Hillary Clinton} instead of \emph{Clinton} or \emph{She}.

\subsection{Leveraging LLMs for Paraphrasing}
To leverage Large Language Models (LLMs) for paraphrasing the RAMS samples, we employed both sentence-level and document-level paraphrasing techniques. Upon examining the examples from Table \ref{t:augmentation_examples_2}, we observed that sentence-level paraphrasing did not facilitate moving the intra-sentential arguments to inter-sentential arguments. This is because we could only provide one sentence as an input to the paraphraser model. However, using the GPT-3 model with the prompt shown in Section \ref{ss:data_augmentation} generated samples with more inter-sentential arguments. In Table \ref{t:augmentation_examples_2},  both intra-sentential arguments (\emph{Clinton and Iran}) from the original sample successfully shifted to inter-sentential arguments. This transition of arguments across sentence boundaries demonstrates the potential of LLMs for enhancing the diversity of training data in natural language processing tasks.

\begin{table*}[h]
\small
\centering
\def\w{.5in}

\begin{tabular}{l p{4.5in}}
\toprule

Original from RAMS
&
As Secretary of State, Hillary Clinton was a hawk on the Iranian nuclear issue. In 2009 - 2010, when Iran first indicated a willingness to compromise, she led the opposition to any negotiated settlement and pushed for punishing sanctions. To clear the route for sanctions, \textcolor{blue}{\argsubg{Clinton}{violator}} helped sink \textcolor{red}{\argsubg{agreements}{event}} tentatively negotiated with \textcolor{blue}{\argsubg{Iran}{otherparticipant}} to ship most of its low - enriched uranium out of the country. In 2009, Iran was refining uranium only to the level of about 3-4 percent, as needed for energy production. Its negotiators offered to swap much of that for nuclear isotopes for medical research.\\ \midrule

Augmented instances \\

~~~Simple Swapping \\
~~~~~~Plain
& 
\textcolor{teal}{\argsubg{Clinton}{violator}} As Secretary of State, Hillary Clinton was a hawk on the Iranian nuclear issue. In 2009 - 2010, when Iran first indicated a willingness to compromise, she led the opposition to any negotiated settlement and pushed for punishing sanctions. To clear the route for sanctions, helped sink \textcolor{red}{\argsubg{agreements}{event}} tentatively negotiated with \textcolor{blue}{\argsubg{Iran}{otherparticipant}} to ship most of its low - enriched uranium out of the country. In 2009, Iran was refining uranium only to the level of about 3-4 percent, as needed for energy production. Its negotiators offered to swap much of that for nuclear isotopes for medical research.\\ \addlinespace

~~~~~~Verbose
&
The violator of the event \textcolor{red}{agreements} is \textcolor{teal}{\argsubg{Clinton}{violator}}. As Secretary of State, Hillary Clinton was a hawk on the Iranian nuclear issue. In 2009 - 2010, when Iran first indicated a willingness to compromise, she led the opposition to any negotiated settlement and pushed for punishing sanctions. To clear the route for sanctions, helped sink \textcolor{red}{\argsubg{agreements}{event}} tentatively negotiated with \textcolor{blue}{\argsubg{Iran}{otherparticipant}} to ship most of its low - enriched uranium out of the country. In 2009, Iran was refining uranium only to the level of about 3-4 percent, as needed for energy production. Its negotiators offered to swap much of that for nuclear isotopes for medical research. \\ \addlinespace \addlinespace

~~~Leveraging Coreference  \\
~~~~~~Random mention
&
 As Secretary of State, Hillary Clinton was a hawk on the Iranian nuclear issue. In 2009 - 2010, when Iran first indicated a willingness to compromise, \textcolor{teal}{\argsubg{she}{violator}} led the opposition to any negotiated settlement and pushed for punishing sanctions. To clear the route for sanctions, Clinton helped sink \textcolor{red}{\argsubg{agreements}{event}} tentatively negotiated with \textcolor{blue}{\argsubg{Iran}{otherparticipant}} to ship most of its low - enriched uranium out of the country. In 2009, Iran was refining uranium only to the level of about 3-4 percent, as needed for energy production. Its negotiators offered to swap much of that for nuclear isotopes for medical research.\\ \addlinespace

~~~~~~Meaningful mention
&
As Secretary of State, \textcolor{teal}{\argsubg{Hillary Clinton}{violator}} was a hawk on the Iranian nuclear issue. In 2009 - 2010, when Iran first indicated a willingness to compromise, she led the opposition to any negotiated settlement and pushed for punishing sanctions. To clear the route for sanctions, Clinton helped sink \textcolor{red}{\argsubg{agreements}{event}} tentatively negotiated with \textcolor{blue}{\argsubg{Iran}{otherparticipant}} to ship most of its low - enriched uranium out of the country. In 2009, Iran was refining uranium only to the level of about 3-4 percent, as needed for energy production. Its negotiators offered to swap much of that for nuclear isotopes for medical research.\\ \bottomrule

\end{tabular}

\caption{Original instance from RAMS and additional instances obtained with our data augmentation strategies (\emph{Simple Swapping} and \emph{Leveraging Coreference}).
See descriptions of the strategies in Section \ref{ss:data_augmentation}).
While \emph{Simple Swapping} results in ungrammatical text as it shuffles intra-sentential arguments, it is the strategy that yields the best results (Table \ref{t:results}).
Leveraging coreference introduces errors when the predicted coreference chains are wrong.}
\label{t:augmentation_examples_1}
\end{table*}

\begin{table*}[h]
\small
\centering
\def\w{.5in}

\begin{tabular}{l p{4.5in}}
\toprule

Original from RAMS
&
As Secretary of State, Hillary Clinton was a hawk on the Iranian nuclear issue. In 2009 - 2010, when Iran first indicated a willingness to compromise, she led the opposition to any negotiated settlement and pushed for punishing sanctions. To clear the route for sanctions, \textcolor{blue}{\argsubg{Clinton}{violator}} helped sink \textcolor{red}{\argsubg{agreements}{event}} tentatively negotiated with \textcolor{blue}{\argsubg{Iran}{otherparticipant}} to ship most of its low - enriched uranium out of the country. In 2009, Iran was refining uranium only to the level of about 3-4 percent, as needed for energy production. Its negotiators offered to swap much of that for nuclear isotopes for medical research.\\ \midrule

Augmented instances \\

~~~Leveraging LLMs \\
~~~~~~Pegasus
&
Hillary Clinton was a strong supporter of the Iranian nuclear issue as Secretary of State. She led the opposition to any negotiated settlement and pushed for punishing sanctions after Iran indicated a willingness to compromise. \textcolor{teal}{\argsubg{Clinton}{violator}} helped sink \textcolor{red}{\argsubg{agreements}{event}} that would have allowed \textcolor{teal}{\argsubg{Iran}{otherparticipant}} to ship most of its low - enriched uranium out of the country. Iran only refined about 3 - 4 percent of its nuclear material in 2009, as needed for energy production. Much of that was offered to be used for medical research. \\ \addlinespace

~~~~~~GPT3 
&
Secretary of State Hillary Clinton has taken a hard line against Iran's nuclear ambitions, recently thwarting \textcolor{red}{\argsubg{agreements}{event}} tentatively negotiated between the two countries. \textcolor{teal}{\argsubg{Clinton}{violator}} has pushed for punishing sanctions, as she argued that any negotiated settlement was not enough to ensure \textcolor{teal}{\argsubg{Iran}{otherparticipant}} would not pursue nuclear weapons. Her opposition to the agreements was based on Iran's offer to swap its low-enriched uranium for nuclear isotopes, which she felt was not sufficient to prevent their potential proliferation. Iran initially sought to refine the uranium to the levels necessary for energy production, but Clinton insisted that the sanctions remain in place. As the situation between Iran and the United States continues to evolve, Clinton's hard-line stance on the Iranian nuclear issue remains. \\ \bottomrule

\end{tabular}

\caption{Original instance from RAMS and additional instances obtained with our data augmentation strategies (\emph{Leveraging LLMs}).
See descriptions of the strategies in (Section \ref{ss:data_augmentation}).}
\label{t:augmentation_examples_2}
\end{table*}

\end{document}